\documentclass[format=sigconf]{acmart}

\usepackage{epsfig}
\usepackage{graphicx}
\usepackage{subcaption}
\usepackage{amsmath}
\usepackage{amssymb}
\usepackage[normalem]{ulem}
\usepackage{multirow}
\usepackage{booktabs} 
\usepackage[font=small]{caption}
\DeclareMathOperator*{\argmax}{argmax}

\graphicspath{ {images/} }

\AtBeginDocument{%
  \providecommand\BibTeX{{%
    \normalfont B\kern-0.5em{\scshape i\kern-0.25em b}\kern-0.8em\TeX}}}

\setcopyright{acmcopyright}
\copyrightyear{2020}
\acmYear{2020}
\acmDOI{10.1145/1122445.1122456}

\acmConference[Seattle '20]{Seattle '20: ACM Multimedia}{October 12--16, 2020}{Seattle, WA}
\acmBooktitle{Seattle '20: ACM Multimedia,
  October 12--16, 2020, Seattle, WA}
\acmPrice{15.00}
\acmISBN{978-1-4503-XXXX-X/18/06}

\begin{document}

\title{Learning Tuple Compatibility for Conditional Outfit Recommendation}

\author{Xuewen Yang}
\email{xuewen.yang@stonybrook.edu}
\affiliation{%
  \institution{WINS Lab, Stony Brook University}
}

\author{Xin Wang}
\email{x.wang@stonybrook.edu}
\affiliation{%
  \institution{WINS Lab, Stony Brook University}
}

\author{Dongliang Xie}
\authornote{Corresponding author}
\email{xiedl@bupt.edu.cn}
\affiliation{%
  \institution{BUPT}
}

\author{Jiangbo Yuan}
\email{yuanjiangbo@gmail.com}
\affiliation{%
  \institution{eBay Inc.}
}

\author{Wanying Ding}
\email{dingwanying0820@gmail.com}
\affiliation{%
  \institution{JPMorgan Chase}
}

\author{Pengyun Yan}
\email{mattpengyunyan@gmail.com}
\affiliation{%
  \institution{Vipshop Inc.}
}

\fancyhead{} 


\begin{abstract}
Outfit recommendation requires the answers of some challenging outfit compatibility questions such as `Which pair of boots and school bag go well with my jeans and sweater?'. 
It is more complicated than conventional similarity search, and needs to consider not only visual aesthetics but also the intrinsic fine-grained and multi-category nature of fashion items.
Some existing approaches solve the problem through sequential models or learning pair-wise distances between items. However, most of them only consider coarse category information in defining fashion compatibility while neglecting the fine-grained category information often desired in practical applications.
To better define the fashion compatibility and more flexibly meet different needs, we propose a novel problem of learning compatibility among multiple tuples (each consisting of an item and category pair), and recommending fashion items following the category choices from customers.
Our contributions include: 1) Designing a Mixed Category Attention Net (MCAN) which integrates both fine-grained and coarse category information into recommendation and learns the compatibility among fashion tuples. MCAN can explicitly and effectively generate diverse and controllable recommendations based on need.
2) Contributing a new dataset IQON, which follows eastern culture and can be used to test the generalization of recommendation systems. 
Our extensive experiments on a reference dataset Polyvore and our dataset IQON demonstrate that our method significantly outperforms state-of-the-art recommendation methods.
\end{abstract} 


\copyrightyear{2020}
\acmYear{2020}
\setcopyright{acmlicensed}\acmConference[MM '20]{Proceedings of the 28th ACM International Conference on Multimedia}{October 12--16, 2020}{Seattle, WA, USA}
\acmBooktitle{Proceedings of the 28th ACM International Conference on Multimedia (MM '20), October 12--16, 2020, Seattle, WA, USA}
\acmPrice{15.00}
\acmDOI{10.1145/3394171.3413936}
\acmISBN{978-1-4503-7988-5/20/10}

\settopmatter{printacmref=true}



\maketitle

\section{Introduction}
Fashion is an essential part of human experiences and has grown into an industry worth hundreds of billions of dollars in US alone \footnote{https://www.statista.com/topics/965/apparel-market-in-the-us/}. With the rapid growth of online shopping, fashion related understanding systems, such as those for outfit recommendation, are now in a great need.
Over the last few years, there has been a remarkable progress in fashion related research, including clothing attribute prediction and landmark detection \cite{Wang2018AttentiveFG, liuLQWTcvpr16DeepFashion}, fashion recommendation \cite{conf/mm/HuYD15, Lin2018ExplainableFR, Yu2018}, clothing item retrieval \cite{Liu:2012, Wang2017ClothingRW}, clothing parsing \cite{2018arXiv180610787G, He2017RealTimeFC} and outfit recommendation \cite{han2017learning, VasilevaECCV18FasionCompatibility, mining16, Lu_2019_CVPR}.

In this paper, we focus on the outfit recommendation problem. An outfit consists of items of different (fine-grained) categories (\textit{e.g.} sweater, jeans, boots, school bag) that have visual aesthetic relationship with one another. 
For outfit recommendation, it requires the learning of \textit{compatibility} \cite{HePM16,VeitKBMBB15,McAuley2015,han2017learning,VasilevaECCV18FasionCompatibility} to show how multiple visual items interact with each other, instead of presenting the \textit{similarity} between items.

\begin{figure}
\centering
\small
\begin{subfigure}[t]{0.15\textwidth}
   \includegraphics[width=1\linewidth]{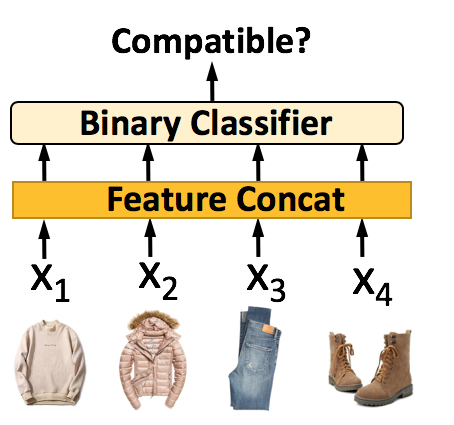}
   \caption{}
   \label{fig:others1} 
\end{subfigure}
\begin{subfigure}[t]{0.15\textwidth}
   \includegraphics[width=1\linewidth]{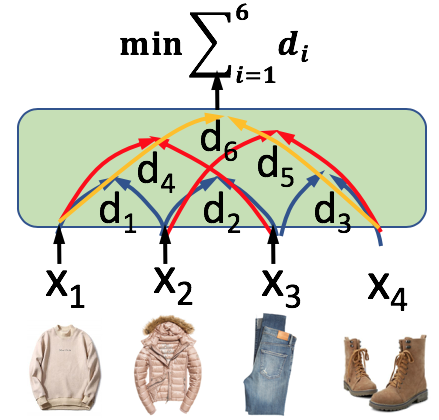}
   \caption{}
   \label{fig:others2}
\end{subfigure}
\begin{subfigure}[t]{0.15\textwidth}
   \includegraphics[width=1\linewidth]{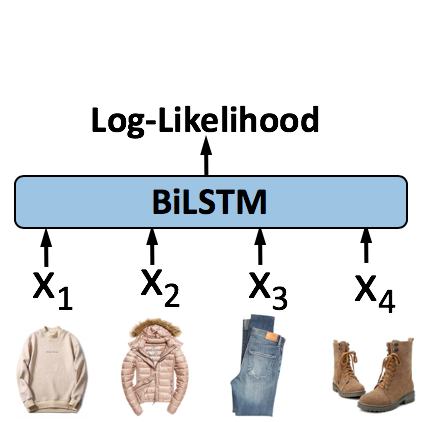}
   \caption{}
   \label{fig:others3}
\end{subfigure}

\caption[]{Compatibility Learning Models: (a) A model training a binary classifier on positive and negative outfits. (b) A model using pair-wise distance as learning metric. 
(c) Sequential models treat an outfit as a sequence and are trained using log-likelihood.}
\label{fig:others}
\vspace{-0.2in}
\end{figure}

Existing studies on outfit recommendation mainly fall into three branches.  The first one~\cite{mining16,McAuley2015} treats an outfit as a set and trains a binary classifier to determine if an outfit is compatible or not based on the concatenation of item features  (Figure~\ref{fig:others1}). As simple concatenation cannot properly model the relationship among items, 
methods of this branch usually perform the worst. 
The second branch~\cite{VeitKBMBB15,HePM16} tries to minimize the sum of distances between every item pairs in the compatibility space   (Figure~\ref{fig:others2}).
This type of methods usually suffers from the \textit{ill-triangle} problem, where two in-compatible items can be considered as compatible when they are both compatible with a third one. 
Although an extended scheme~\cite{VasilevaECCV18FasionCompatibility} adds the `type' information to the compatibility learning, it needs to learn the `type' embedding matrices for every type pair, which is very inefficient as the time complexity is $\mathcal{O}(n^2)$ for $n$ types of clothes.
The third branch~\cite{han2017learning,Takuma} considers an outfit as a sequence of items in a pre-defined order (Figure~\ref{fig:others3}). It models the compatibility of an outfit using LSTM, which is `order-dependent'. When the order of the outfit varies, the performance greatly degrades.

As another limitation, existing recommendation systems tend to consider only visual aesthetics and meta data such as item descriptions. Although these meta data can be used for describing image features, they are not for compatibility learning. Coarse category information (\textit{e.g.} tops, bottoms, shoes, etc.) is also applied to define an outfit as a sequence~\cite{han2017learning,Takuma}, where an item of a pre-defined coarse category is provided at each time step. 
Training with only coarse or fine category but not both is not flexible because the category of a query item for inference can be coarse while the model is trained with fine categories. There is a lack of consideration of both category classes in the literature work.
To meet the need of customers, both fine-grained and coarse categories should be considered. For example, two customers with the same T-shirt may be interested in different types of pants, such as jeans or sweat pants, rather than simply a `bottom'. In another case, a customer with a T-shirt wants to have a `bottom' of any type.  To incorporate both fine and coarse category information into the compatibility learning and make the outfit recommendation fit for the flexible need, we introduce the notion \textit{tuple}. A tuple is composed of a fashion item and its category, which can be either coarse or fine. Accordingly, we extend the learning of item compatibility to the learning of tuple compatibility.

To enable the learning of tuple compatibility and address issues existing in the literature as discussed above, in this work, we propose a Mixed Category Attention Net (MCAN) which considers an outfit as a combination of tuples formed with item-category pairs and learns the compatibility among them. Given a partial tuple sequence, MCAN can predict the next item through the  \textit{item prediction layer} (IPL) embedded in MCAN. The item will fall into the category required if the next category is given; otherwise, the next category will be first predicted using a \textit{category prediction layer} (CPL). 
Incorporating the sequence of categories into the input, MCAN can be much more versatile in recommendation, and can support a set of applications.
First, MCAN can be trained with data of different category types to provide accurate recommendation, regardless that an item is given with a fine or coarse category. Second, the items in an outfit can be `order-free', with the order of items following that of categories provided at a time, rather than being fixed as in sequential models.
Third, introducing `category' into the compatibility learning helps to reduce the searching space of recommendation, which not only helps to greatly reduce the searching complexity but also helps to significantly increase the accuracy in recommendation. Fourth, MCAN can support the `filling in the blank' application, where missing items from the designated categories can be recommended to form a complete outfit, according to partial items provided by customers.
By exposing to more training orders, MCAN is more explicit and effective in generating diverse recommendation results than the methods~\cite{han2017learning,Takuma, VasilevaECCV18FasionCompatibility} that apply beam search or swapping of similar items.
MCAN exploits \textit{self-attention} to learn representations for compatibility learning. 
Through extensive experimental studies, we find that MCAN is efficient in learning both \textit{one-one} (item-item) and \textit{one-many} (item with partial outfit) dependency, owing to its capability of learning dependency among tuples.  
This allows it to  avoid the \textit{ill-triangle} problem existing in methods using only the pair-wise (\textit{one-one}) distance learning in the metric.

As another important contribution, since the public available fashion datasets have a lot of noisy items \cite{han2017learning,Tangseng2017, VasilevaECCV18FasionCompatibility}, we generate a new dataset IQON with detailed category labels, as described in Section~\ref{dataset}. Complementary to the existing datasets based on the western culture, it follows the eastern culture, which helps  to evaluate the generalization of recommendation systems by running over different types of datasets. 
The contributions of this paper are:
\begin{itemize}
\itemsep=0em
\item We introduce a novel concept of tuple compatibility learning to enable more accurate and flexible outfit recommendation.
\item We propose a MCAN model that can flexibly recommend one or several missing items from an outfit according to customers' preferences without restricting the order of items or the category class of an item. The search can follow any order based on the sequence of item categories given or inferred, and an item can be identified with either a fine or a coarse category.
\item We collect a large dataset \textit{IQON} that can facilitate the research related to fashion compatibility learning, where the  data are of different categories and outfit styles. This dataset will be released for future research by the community.
\end{itemize}


\section{Related Work}
\label{related}

\textbf{Fashion Compatibility Learning}
Some fashion recommendation methods \cite{HePM16,Cong2018,Lin2019,VasilevaECCV18FasionCompatibility,VeitKBMBB15,Yin2019} model the compatibility between item pairs by minimizing the total pair-wise distances in the latent compatibility space.
These methods are prone to the \textit{ill triangle} problem.
Composition recommendation method \cite{mining16,McAuley2015} treats an outfit as a composition with its feature formed by concatenating all item features. Then a binary classifier is trained to distinguish compatible outfits from incompatible ones. However, simple concatenation cannot model the complex dependency among outfit items. Sequential recommendation methods \cite{han2017learning,Takuma} treat an outfit as a prefixed sequence of coarse categories-top, bottoms, shoes, and bags. It finds the item compatibility by learning the transition between time steps using an LSTM model, which will suffer from big performance degradation if the order changes.

Different from existing work, we propose MCAN to model the compatibility among \textit{tuples}, with each tuple formed with an item and category pair. Taking advantage of the order-independent feature of \textit{self-attention},
the category prediction layer (CPL) embedded in MCAN can help to provide the order of categories.
The items to recommend can directly follow the order of categories provided by customers, or the order inferred by the CPL layer if the order is not provided. The introduction of the sequence of categories not only reduce the search space, but helps to increase the accuracy of recommendation while providing customers the flexibility to meet different  needs through the sequence and categories given.
As another benefit, MCAN can enable parallel training to improve the computational efficiency.


\section{Datasets}
\label{dataset}
To evaluate the efficiency and effectiveness of our model, we test it 
using the following two datasets:

\textbf{Polyvore Dataset}: \textit{\url{polyvore.com}} is a fashion website where users could create, share and comment fashion outfits online. The Polyvore dataset provides rich information such as images, text descriptions, tags, number of likes, and category information. Han \textit{et al.}  \cite{han2017learning} supplied a dataset using Polyvore outfits that has a lot of irrelevant images,  such as logo images, decoration images, background images, blank images, and repeated images. 
We remove all irrelevant ones and keep only the images related to fashion. This dataset still has three main drawbacks. First, it mainly contains luxury clothing brands which cannot satisfy people from different classes. Second, it is relatively small with only 17316 outfits after the filtering. Third, some outfits that only few people like. 

\textbf{IQON Dataset}: To enable the recommendation with fine grained categories and design a larger dataset, we create a new dataset IQON. We collect the outfits from the Japanese fashion website  \textit{\url{iqon.jp}}, which allows users to create their own outfits by collecting  images of fashion items. The items are from both luxury clothing brands and economic ones. We collect over 400000 outfits with over 1 million items and keep the outfits that more than 70 people like, which reduces the dataset to contain only 28883 outfits. Each item has a fine-grained category, such as blouser and sweater. We use a well trained classifier to make sure that items in the same outfit are from different categories. This greatly helps to remove noisy items in the training data. We split the IQON dataset into training (20000 outfits), testing (4400 outfits), and validation (4483 outfits).

Compared with the previous datasets~\cite{han2017learning,Tangseng2017, VasilevaECCV18FasionCompatibility} collected from \textit{\url{polyvore.com}} fashion website, our dataset is a good benchmark for evaluating the generalization  of recommendation algorithms for fashion items from a different culture. Moreover, our dataset contains outfits only preferred by more than 70 people, while the other datasets do not filter out the unpopular outfits that can affect the performance of the algorithms.

\section{Learning Tuple Compatibility with Mixed Category Attention Net}
\label{sec:model}
In this section, we first define our tuple compatibility learning problem. We then present in details the design of a Mixed Category Attention Net (MCAN) that can flexibly recommend items following the sequence of fine or coarse categories.
We will further propose a \textit{tuple triplet learning} scheme to improve the performance of MCAN. 
With the learning of compatibility among tuples, 
MCAN is able to recommend missing items of specific categories when given an incomplete outfit. 
The order of items to recommend can be determined by the sequence of fine or coarse categories, which can be either provided by customers according to their preferences or generated by MCAN.
Finally, we introduce its application in conditional fashion recommendation.

\subsection{Learning of Tuple Compatibility}
An outfit is a collection of items of different categories that have visual aesthetic relationship with one another. 
We consider both fine-grained and coarse category information. 
Taking the category into consideration, an outfit can be defined as a sequence of tuples $O=\{ (\mathbf{x}_1, \mathbf{c}_1), \ldots, (\mathbf{x}_i, \mathbf{c}_i), \ldots, (\mathbf{x}_N, \mathbf{c}_N) \}$, with $N$ being the number of items in an outfit, and $\mathbf{x}_i$ being the embedding of the fashion item from the category $\mathbf{c}_i$.\footnote{Through out this paper, we use notion of category and that of category embeddings inter-changeably.} 
Although a sequence is given in $O$, our unique design makes our model \textit{order-free}, with the order of items simply following that of the categories given as input.

An outfit will be observed more frequently if it is more compatible. Therefore, we use the joint distribution to measure the compatibility of an outfit, which is defined as:
\begin{equation}
\small
    \mathcal{C} = P(\mathbf{x}_1, \mathbf{c}_1)\prod_{i=1}^{N-1}P\big((\mathbf{x}_{i+1}, \mathbf{c}_{i+1})\vert (\mathbf{x}_1, \mathbf{c}_1), \ldots, (\mathbf{x}_{i}, \mathbf{c}_{i})\big)
\label{eq:com}
\end{equation}
The first tuple $(\mathbf{x}_1, \mathbf{c}_1)$ is usually provided by the customers.

For an outfit with a set of tuples already determined, this definition also allows the learning of the compatibility between a new tuple and existing ones. 
To reduce the searching space of Eq.~\ref{eq:com}, if the category for an item to select is not given, we can first search for the best category and then try to find the item with the highest probability of the category. This procedure can be written by first translating the Eq.~\ref{eq:com} using the chain rule and then taking a logarithm:
\begin{equation}
\begin{aligned}
    \mathcal{C}
    &=\sum_{i=1}^{N-1}\big(\log P\big(\mathbf{x}_{i+1}\vert (\mathbf{x}_1, \mathbf{c}_1), \ldots, (\mathbf{x}_{i}, \mathbf{c}_{i}), \mathbf{c}_{i+1}\big) \\
    &+\log P\big(\mathbf{c}_{i+1}\vert (\mathbf{x}_1, \mathbf{c}_1), \ldots, (\mathbf{x}_{i}, \mathbf{c}_{i})\big)\big) + \log P(\mathbf{x}_1, \mathbf{c}_1)
\end{aligned}
\label{eq:com_chain_log}
\end{equation}

\subsection{Mixed Category Attention Net}
\label{fccan}
\begin{figure}
\centering
\includegraphics[width=0.4\textwidth]{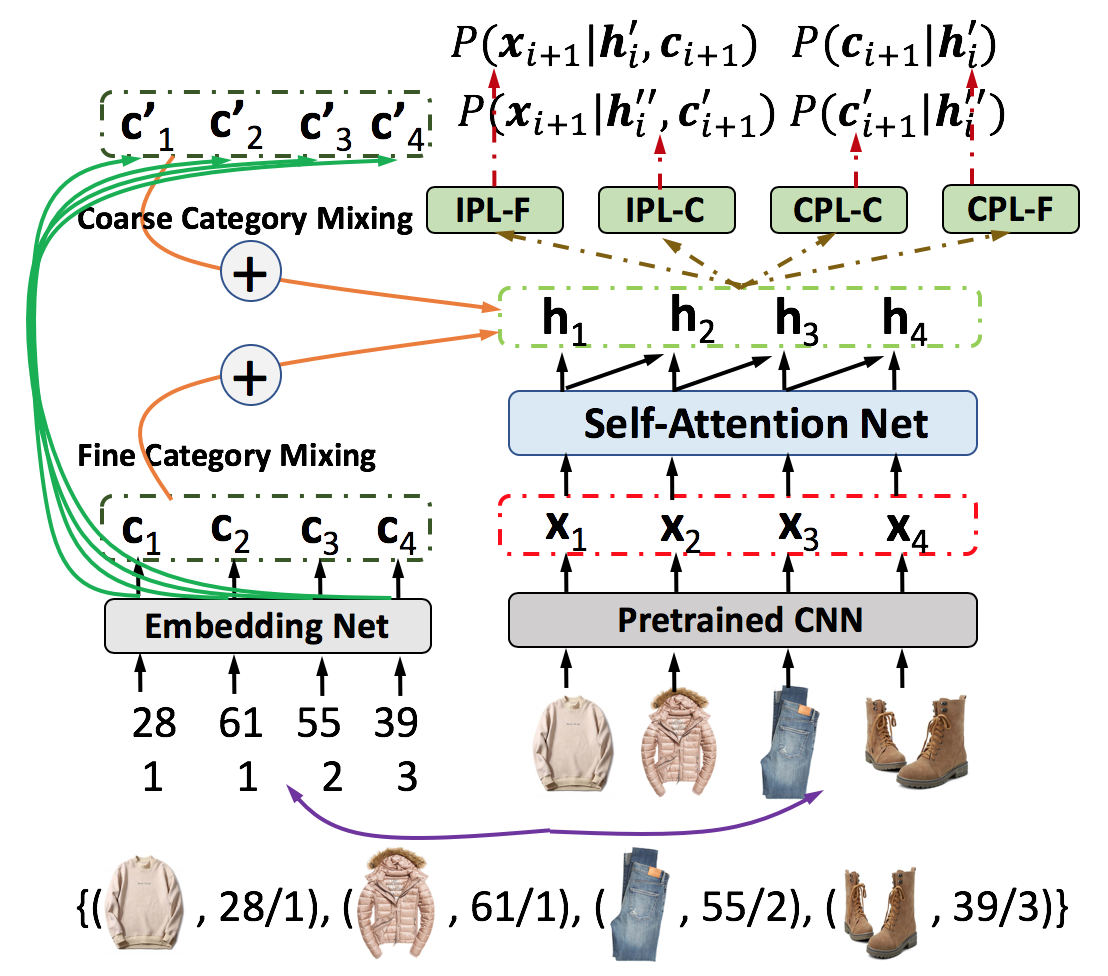}
\caption{MCAN takes a sequence of tuples as input. The Embedding Net encodes both the fine ($28, 61, 55, 39$) and coarse ($1,1,2,3$) categories into \textit{category representations}. The SAN takes the \textit{category representations} and features extracted from the input outfit items to align the output items with the category sequence. Category Prediction Layers (CPL-F for fine categories and CPL-C for coarse categories) output the probability of the next category given the previous tuple sequence. Item Prediction Layers (IPL-F for fine categories and IPL-C for coarse categories) output the probability of the next item given previous tuple sequence and the next category.}
\label{fig:model}
\vspace{-0.15in}
\end{figure}

To model the compatibility of an outfit defined by Eq.~\ref{eq:com_chain_log} using both fine and coarse category information, we propose a Mixed Category Attention Net (MCAN). It allows the category to be either fine or coarse, and takes advantage of the self-attention net (SAN) to exploit the dependency among items in an outfit without fixing their order.
As shown in Figure~\ref{fig:model}, MCAN consists of two main components, an Embedding Net (EN) and an SAN. With a sequence of tuples as input, MCAN sends the sequence of items and the sequence of categories to a pretrained CNN and Embedding Net respectively to extract the image features $\{ \mathbf{x}_1,\ldots,\mathbf{x}_N\}$ and category features $\{ \mathbf{c}_1,\ldots,\mathbf{c}_N\}$.

To obtain the features of the tuples, the image features $\mathbf{X}=\{ \mathbf{x}_1,\ldots,\mathbf{x}_N \}\in \mathbb{R}^{N\times d}$ ($d$ being the feature dimension) are first transformed into two matrices through linear projections, $\mathbf{F}(\mathbf{X})=\mathbf{W}_{\mathbf{F}}\mathbf{X}$ and $\mathbf{G}(\mathbf{X})=\mathbf{W}_{\mathbf{G}}\mathbf{X}$, which are applied to find the attention as follows:
\begin{equation}
    \alpha_{ij}=\frac{exp(e_{ij})}{\sum_{k=1}^Nexp(e_{ik})}
\end{equation}
where $e_{ik}=a(\mathbf{F}_i, \mathbf{G}_k$). $\alpha_{ij}$ indicates the extent to which the model attends to item $i$ when recommending item $j$. $a$ is a feed-forward neural network which is jointly trained with other components of the model. 
In the above formulation, $\mathbf{W}_{\mathbf{F}},\mathbf{W}_{\mathbf{G}}$ are learned weight matrices, which are implemented as $1\times 1$ convolutions.
The output of the SAN is $\mathbf{H}=\{ \mathbf{h}_1,\ldots,\mathbf{h}_j,\ldots,\mathbf{h}_N \}\in \mathbb{R}^{N\times d}$.
To model the compatibility among tuples defined by Eq.~\ref{eq:com_chain_log}, we first propose a \textit{Feature Mixing} operation to pair the representations of item sequence $\{ \mathbf{h}_1,\ldots,\mathbf{h}_N\}$ with category sequence $\{ \mathbf{c}_1,\ldots,\mathbf{c}_N\}$ to form the tuple representation:   
\begin{equation}
    \mathbf{t}_i=f([\mathbf{h}_i, \mathbf{c}_i]),
\end{equation}
where $f$ is a feed-forward network, $[\mathbf{h}_i, \mathbf{c}_i]$ is a feature concatenation, and the dimension of $\mathbf{t}_i$ is the same as that of $\mathbf{h}_i$.
The tuple representation $\mathbf{t}_i$ is then fed into the \textit{Item Prediction Layer (IPL)} in Figure~\ref{fig:model} to calculate the probability of the next item given previous tuples: 
\begin{equation}
\begin{aligned}
    P&\big(\mathbf{x}_{i+1}\vert(\mathbf{x}_1, \mathbf{c}_1), \ldots, (\mathbf{x}_{i}, \mathbf{c}_{i}),\mathbf{c}_{i+1}\big)
    =P(\mathbf{x}_{i+1}\vert \mathbf{t}_{1}, \ldots,\mathbf{t}_{i},\mathbf{c}_{i+1})\\
    &=\frac{exp(s(\mathbf{t}_{i}, f([\mathbf{W}_{\mathbf{H}}\mathbf{x}_{i+1}, \mathbf{c}_{i+1}])))}{\sum_{\mathbf{x}\in \mathcal{X}}exp(s(\mathbf{t}_{i}, f([\mathbf{W}_{\mathbf{H}}\mathbf{x}, \mathbf{c}_{i+1}])))},
\end{aligned}
\label{eq:prob}
\end{equation}
where $s$ is a scoring function, modeled by a feed-forward network, to signify the relationship between the next tuple and the previous ones.
To constrain the predicted item to belong to the category $\mathbf{c}_{i+1}$, we implement IPL using a \textit{softmax layer}, where $\mathcal{X}$ is the subset formed by all items belonging to the category $\mathbf{c}_{i+1}$.  The items not belonging to this category will output the probability 0. The identification of the category helps improve the accuracy, while restricting the items to a subset on the softmax layer reduces the computational complexity.

Similarly, the probability that the next recommended item falls into the category $\mathbf{c}_{i+1}$ is defined as:
\begin{equation}
    P\big(\mathbf{c}_{i+1}\vert(\mathbf{x}_1, \mathbf{c}_1), \ldots, (\mathbf{x}_{i}, \mathbf{c}_{i})\big)=
    P(\mathbf{c}_{i+1}\vert \mathbf{t}_{1}, \ldots,\mathbf{t}_{i})
\label{eq:prob_c}
\end{equation}
where $\mathbf{c}_{i}\in C$ is the  set that contains all categories in the study. This is modeled by a \textit{Category Prediction Layer} (CPL, a softmax layer over all categories) on Figure~\ref{fig:model}.

Thus, the training loss function of learning using fine categories can be defined as:
\begin{equation}
\small
\begin{aligned}
    \mathcal{L}_{F}=&-\sum_{i=1}^{N-1}\big(\log P(\mathbf{x}_{i+1}\vert \mathbf{t}_{i},\mathbf{c}_{i+1})+\log P(\mathbf{c}_{i+1}\vert\mathbf{t}_{i})\big)
\end{aligned}
\label{eq:san}
\end{equation}

In real-world applications, there are scenarios where exact fine-grained categories are not provided or hard to determine. For example, an item with a new style may not fall into any known fine category, but can easily be identified as `a top'. To make MCAN more flexible and work when only coarse categories are provided, we jointly train it with the information from both fine and coarse categories.
Similar to the case for fine categories, we can represent a tuple $i$ with the coarse ones included:
\begin{equation}
    \mathbf{t}_i^{\prime}=f^{\prime}([\mathbf{h}_i, \mathbf{c}^{\prime}_i])
\end{equation}
with $f^{\prime}$ being a feed-forward network and $\mathbf{c}^{\prime}_i$ being coarse category embedding for item $i$. We can then define the training loss function for learning with coarse categories as:
\begin{equation}
\small
\begin{aligned}
    \mathcal{L}_{C}=&-\sum_{i=1}^{N-1}\big(\log P\big(\mathbf{x}_{i+1}\vert \mathbf{t}_i^{\prime},\mathbf{c}^{\prime}_{i+1}\big)+\log P(\mathbf{c}^{\prime}_{i+1}\vert\mathbf{t}_{i}^{\prime})\big)
\end{aligned}
\label{eq:san-c}
\end{equation}

\subsection{Tightening the Compatibility Space with Triplet Learning for Tuples}
\label{sec:triplet}
\begin{figure}
\centering
\includegraphics[width=0.4\textwidth]{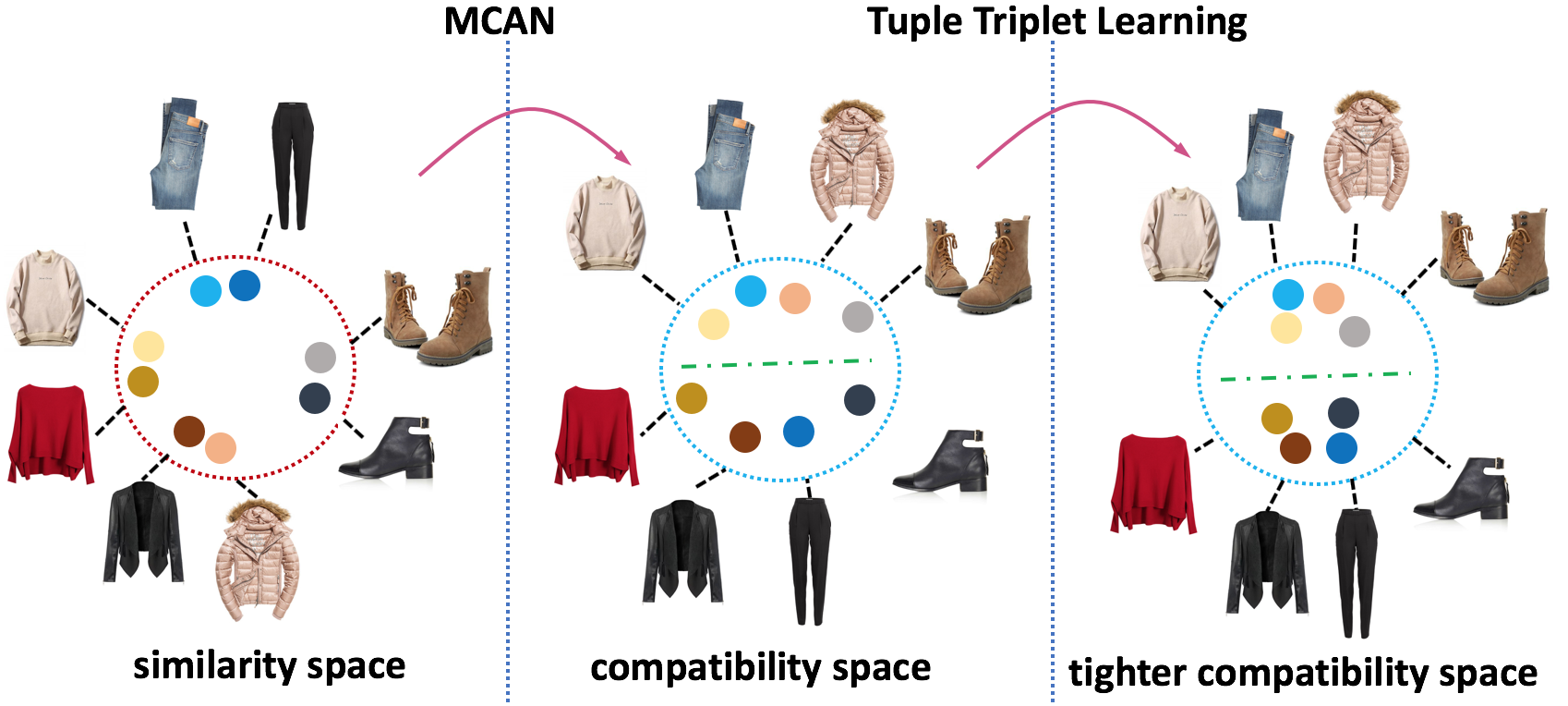}
\caption{Items of the same category (e.g., boots) are close to each other in the similarity space (on the left of the figure). With MCAN (middle of the image), they are separated by the green dashed line into different compatible sets.
With the incorporation of \textit{tuple triplet training}, the compatible tuples are put even closer to each other but farther away from incompatible ones, making the compatibility space tighter.}
\label{fig:an-triplet}
\vspace{-0.15in}
\end{figure}

MCAN is expected to separate an outfit from another negative one by some un-quantifiable compatibility margin, as shown in the middle of Figure ~\ref{fig:an-triplet}. 
It is hard to set a reasonable margin for the distance between the positives and the negatives. If this margin is small, two outfits may be still close, so the items of one outfit are not well separated from the items of the other outfit. To further reduce the distances between items of the same outfit while increasing the distances from other ones, we propose the use of \textit{tuple-based triplet training} to increase the compatibility of items belonging to the same outfit.

Different from previous triplet definition~\cite{SchroffKP15, VasilevaECCV18FasionCompatibility}, where
a triplet is defined as a set of three items, we define it as a partial sequence of tuples $\{(\mathbf{x}_1, \mathbf{c}_1),\ldots,(\mathbf{x}_i,\mathbf{c}_i)\}$ (anchor), a target tuple $(\mathbf{x}^p_{i+1}, \mathbf{c}_{i+1})$ (positive) and a negative tuple $(\mathbf{x}^n_{i+1}, \mathbf{c}_{i+1})$ (negative), annotated as $\{ \mathbf{t}^a_{i}, \mathbf{t}^p_{i+1}, \mathbf{t}^n_{i+1} \}$.
The anchor $\mathbf{t}^a_i$ is considered to be compatible with the positive $\mathbf{t}^p_{i+1}$ if items of its partial outfit appear also in an outfit with the positive tuple, and to be incompatible with the negative $\mathbf{t}^n_{i+1}$ if it does not appear in any outfit with the negative one. Both positive and negative items selected belong to the same category $\mathbf{c}_{i+1}$. The outfit triplet loss for the whole outfit is then written as
\begin{equation}
\mathcal{L}_{triplet}=\sum_{i=1}^{N-1}\big(s(\mathbf{t}_i^a,\mathbf{t}_i^{n+1})-s(\mathbf{t}_i^a,\mathbf{t}_{i+1}^p)+\mu \big)
\label{loss_tri}
\end{equation}
where $\mu$ is a margin value, $s$ is the scoring function that signifies the relationship between the next tuple and tuples already selected for an outfit, the same as Eq.~\ref{eq:prob}.

Together with the losses for learning with fine categories $\mathcal{L}_{F}$ (Eq.~\ref{eq:san}) and coarse categories $\mathcal{L}_{C}$ (Eq.~\ref{eq:san-c}), our MCAN model minimizes the total loss:
\begin{equation}
\mathcal{L} = \mathcal{L}_{F} + \lambda_1\mathcal{L}_{C} +  \lambda_2\mathcal{L}_{triplet}
\end{equation}
where $\lambda_1$ and $\lambda_2$ are tunable parameters.

\vspace{0.06in}
\noindent{{\bf Negative Sampling Method.}} The tuple-based triplet training requires the sampling of negative tuples.
In the conventional triplet learning, the negative items are the ones that are not in the same set as the target one~\cite{han2017learning, VasilevaECCV18FasionCompatibility,Takuma,tanSimilarity2019}.
To be consistent with this definition, we consider a {\em negative tuple} of a target as the one that contains a negative item but belongs to the same category.
We denote the negative sampling without considering the category information as the \textit{easy} negative sampling.
We further introduce two new negative sampling methods as follows.

\textbf{{\em Semi-Hard Negative Sampling:}}
Semi-hard negative samples are tuples that fall into  the same coarse category as the target one but are from different outfits. This helps our model to learn to incorporate the coarse category information and distinguish a positive item from semi-hard negative ones.

\textbf{\em Hard Negative Sampling:}
Hard negative samples are taken from the same fine category as the target one, but from different outfits. 
This helps MCAN correctly integrate the fine category information.

In our experiments, we use \textit{semi-hard} negative sampling for the first half of the training time and \textit{hard} negative sampling for the second half of the training time, gradually tightening the compatibility space.

\subsection{Conditional Recommendation with MCAN}
\begin{figure}
\centering
\includegraphics[width=0.4\textwidth]{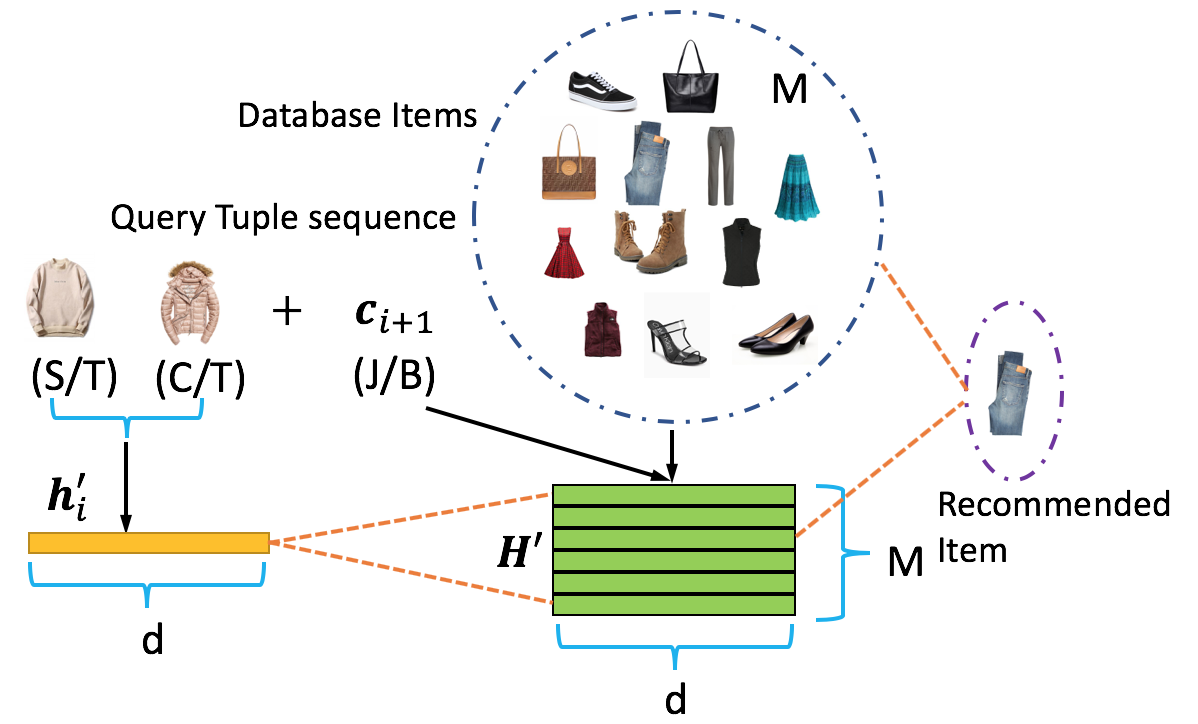}
\caption{Given the previous $i=2$ tuples and the category of target item (fine or coarse) $\mathbf{c}_{i+1}$, our model selects the most compatible item from the database. Abbreviations: S (sweater), C (coat), J (jeans), T (top), B (bottom).}
\label{fig:prediction}
\vspace{-0.15in}
\end{figure}

We give an example to illustrate how the MCAN can be used in a recommendation process using Figure~\ref{fig:prediction}. We demonstrate this using tuples formed by fine categories. For scenarios where only coarse categories are provided, the recommendation process is similar.

In the first scenario, the tuples with fine-grained categories are provided by the customer.
Then our model uses Eq.~\ref{eq:prob} to select the most possible item from database. In Figure~\ref{fig:prediction}, suppose a customer already has a sweater and a coat and wants to make a complete outfit with a jean (category $\mathbf{c}_{i+1}$). MCAN first maps the query tuple sequence to get $\mathbf{h}_i^{\prime}$. Then, every item from the target category in the database will be bundled with $\mathbf{c}_{i+1}$ to form a tuple. Thus our model get mixed features for each new tuple to form a matrix $\mathbf{H^{\prime}}$, with row $j$ stands for a tuple feature of the $j$-th item in the database. Finally, the one that yields the highest probability with Eq.~\ref{eq:prob} will be selected as the next item.

In the second scenario, only partial outfit is given but not the category of the next item to recommend. MCAN can use Eq.~\ref{eq:prob_c} to first recommend the most likely category for the next item and then use Eq.~\ref{eq:prob} to select the item of this category from the database.

As the recommended item aligns with the category provided,  by feeding different categories, our model is able to provide controllable recommendation. Compared to existing models that use beam search or similar item swapping, 
the prediction of different category sequences through the embedded CPL allows recommending diverse outfits  more explicitly and effectively. We demonstrate this benefit in  Section~\ref{sec:cdr}.

\section{Experiments}
\label{experiments}
We conduct experiments over two datasets \textbf{Polyvore Dataset} and our collected \textbf{IQON Dataset}
using the following two tasks:
\begin{itemize}
\itemsep=0em
\item \textbf{Outfit compatibility prediction}: Score an outfit to be compatible or not. 
\item \textbf{Fill-in-the-blank (FITB) outfit recommendation}: For a given partial outfit, one has to select the item from a set of candidates that is most compatible with it. 
As a simple example, a user may want to choose pants to match his T-shirt and shoes. 
\end{itemize}

For all tasks, we use \textit{SGD} optimizer for training, with the initial learning rate set to 0.01 and the batch size set to 20. After testing with different values, the hyper parameters are set as $\mu=0.05$, and $\lambda_1=\lambda_2=0.1$. The dimension of the compatibility space is $d=512$.
A CNN is pre-trained  using ImageNet, following Inception-v3~\cite{Szegedy16}. 
We use an Embedding Net to map a category to a dense vector, where the Embedding Net is a look-up table of features for the categories.
The $i$-th row corresponds to the embedding of the $i$-th category. The look-up table is initialized randomly and then updated in each iteration of the training.
Self-Attention Net (SAN) is a network to calculate high-level feature representations of images.

\subsection{Negative Sampling at Different Levels of Difficulty}
\label{sec:ns2}
When testing the performance of different models, existing methods \cite{han2017learning, Takuma} create test sets with \textit{easy} negative sampling, which we denote as easy tasks.  However, this could lead an outfit to contain items of the same category. 
Many of these samples can be easily removed without even considering the item compatibility. 
For example, an outfit with multiple T-shirts can be easily identified as non-compatible.
To overcome this drawback and focus on outfits that cannot be filtered out in
an easy manner, we 
apply our \textit{semi-hard} and \textit{hard} negative sampling to create two extra tasks in addition to the \textit{easy} one.
Specifically, we evaluate our model on creating tasks of three difficulty levels:
\begin{itemize}
    \item \textit{easy}: We use the \textit{easy} negative sampling methods to get negative samples to create testsets, the same as those in~\cite{han2017learning, VasilevaECCV18FasionCompatibility,Takuma,tanSimilarity2019}. For Polyvore dataset, we directly use the one in~\cite{han2017learning}. 
    While for IQON dataset, we create a new one.
    \item \textit{semi-hard} and \textit{hard}: We create one testset on Polyvore, one on IQON using semi-hard negative sampling.
    Similarly, a hard testset is created using hard negative sampling.
\end{itemize}

\subsection{Reference Approaches}
 MCAN is trained on tuples with both fine and coarse categories. In the test phase, we evaluate its performance in two scenarios: outfits with fine categories and outfits with only coarse categories. This makes two comparison cases: \textit{MCAN (coarse)} and \textit{MCAN} (fine). Although our model can work in either case, as peer algorithms only work in one of the cases, we compare all models with only one category information.
 We further add our \textit{tuple triplet learning} scheme to our training method and introduce two other cases to evaluate: \textit{MCAN + Triplet (coarse)} and \textit{MCAN + Triplet (fine)}.

In order to evaluate the functionality of the \textit{Category Prediction Layer} (CPL), we remove it and see how it affects the performance of MCAN. We denote the new model as MAN. Similarly, we can have two comparison cases: MAN (coarse) and MAN (fine).

We compare the above variants of MCAN with the following state-of-the-art approaches:

\textbf{Siamese Net} \cite{VeitKBMBB15} uses a Siamese CNN to map item images to a latent space that expresses compatibility.
No category information is added. Thus, we denote it as `Siamese Net (none)'.

\textbf{BiLSTM} \cite{han2017learning} Compatibility learning is achieved by learning the transition between items of different time steps. Coarse categories are implicitly added by pre-defining items of different time steps to be of specific coarse categories. Thus, we denote this model as `BiLSTM (coarse)'.

\textbf{CSN}~\cite{VasilevaECCV18FasionCompatibility} extends Siamese Net~\cite{VeitKBMBB15} by introducing fine categories. It maps the image features extracted from Siamese Net to `type-aware' compatibility space by using pair-wise mapping matrices conducted for every type pair. As directly providing coarse category information in test phase to the model trained with fine ones is impossible, we re-train it from scratch with coarse category information and denote the two as `CSN (coarse)' and `CSN (fine)'.

\textbf{FHN}~\cite{Lu_2019_CVPR} learns binary codes for fashion outfits compatibility using a set of type-dependent hashing modules.  For fair comparison, we use the objective with item-item pairs, the same as CSN.
As the model architecture of FHN is very similar to that of CSN, we use the same method to train `FHN (coarse)' and `FHN (fine)'.

\begin{table*}[!t]
\caption{Comparison of different methods on IQON Dataset on the FITB and compatibility prediction tasks, with tasks created using easy, semi-hard and hard negative sampling methods. Results are provided for both ordered/unordered cases on the left and right of `/'. }
\small
\centering
\begin{tabular}{cccccccl}
\toprule
\multirow{3}{*}{Method} & \multicolumn{3}{c}{FITB accuracy (\%)} & \multicolumn{3}{c}{Compat. AUC} \\\cline{2-7}
& easy & semi-hard & hard & easy & semi-hard & hard \\
\midrule
Siamese Net \cite{VeitKBMBB15} (none) & 53.0/53.0 & 47.8/47.8 & 43.4/43.4 & 0.80/0.80 & 0.75/0.75 & 0.71/0.71 \\
Bi-LSTM \cite{han2017learning} (coarse) & 70.4/62.2 & 67.3/60.5 & 64.8/57.3 & 0.84/0.74 & 0.80/0.71 & 0.72/0.66 \\
FHN~\cite{Lu_2019_CVPR} (coarse) & 68.5/68.5 & 65.7/65.7 & 63.0/63.0 & 0.82/0.82 & 0.77/0.77 & 0.71/0.71 \\
FHN~\cite{Lu_2019_CVPR} (fine) & 78.5/78.5 & 74.8/74.8 & 69.3/69.3 & 0.87/0.87 & 0.80/0.80 & 0.73/0.73 \\
CSN \cite{VasilevaECCV18FasionCompatibility} (coarse) & 69.1/69.1 & 66.6/66.6 & 64.2/64.2 & 0.84/0.84 & 0.79/0.79 & 0.71/0.71  \\
CSN \cite{VasilevaECCV18FasionCompatibility} (fine) & 80.3/80.3 & 76.9/76.9 & 72.8/72.8 & 0.90/0.90 & 0.85/0.85 & 0.76/0.76  \\
\hline
MAN (coarse) & 78.1/77.4 & 76.8/73.9 & 73.7/70.1 & 0.89/0.85 & 0.85/0.82 & 0.80/0.75 \\
MAN (fine) & 83.6/80.8 & 80.2/78.6 & 78.1/75.2 & 0.92/0.87 & 0.89/0.83 & 0.82/0.76\\
\hline
MCAN (coarse) & 79.4/79.3 & 77.6/77.4 & 75.3/75.3 & 0.91/0.91 & 0.87/0.86 & 0.81/0.80 \\
MCAN (fine) & 85.1/85.0 & 82.8/82.7 & 79.5/79.3 & 0.95/0.94 & 0.91/0.90 & 0.85/0.83\\
MCAN+ Triplet (coarse) & 81.2/81.2 & 79.3/79.2 & 77.7/77.6 & 0.94/0.93 & 0.89/0.89 & 0.83/0.82 \\
MCAN + Triplet (fine)  & \textbf{86.5}/\textbf{86.4} & \textbf{84.1}/\textbf{84.0} & \textbf{82.0}/\textbf{81.8} & \textbf{0.96}/\textbf{0.95} & \textbf{0.93}/\textbf{0.93} & \textbf{0.88}/\textbf{0.87} \\
\bottomrule
\end{tabular}
\label{tab:iqon}
\end{table*}

\subsection{Prediction of Outfit Compatibility}
We define the compatibility of our proposed MCAN using Eq.~\ref{eq:com_chain_log}.  Complementary to the test dataset generated by~\cite{han2017learning} which uses easy negative sampling, we also create two other test datasets that use semi-hard negative sampling and hard negative sampling to choose negative samples. This helps to compare the ability of different models in integrating coarse and fine category information  to correctly identify negative items and filter them out during outfit recommendation. We evaluate the performance using AUC of the ROC curve. Results on IQON and Polyvore datasets are shown in the right part of Table~\ref{tab:iqon} and~\ref {tab:polyvore} respectively. 

Easy negative set may contain items from the same category of the ones already existing in partial outfits that can be easily identified. As it is harder for models to filter out the negative items from the same coarse or fine category of the target item, the performances of all models drop for semi-hard and hard cases. However, the relative improvement of MCAN over other models increases as the task becomes harder. Compared to CSN (fine), in terms of compatibility AUC, MCAN improves $6.0\%$, $8.0\%$ and  $12.0\%$ in easy, semi-hard and hard tasks. 
This demonstrate the capability of MCAN in effectively incorporating the category information to increase the prediction accuracy.

On all tasks of both datasets, Siamese Net (none) achieves the worst results compared with other models. The main reason is that it only utilizes item features without considering any category information. Bi-LSTM (coarse), FHN (coarse) and CSN (coarse) perform better than Siamese Net (none) because they implicitly encode coarse category information in the model, either by forcing a pre-defined sequence order (order encoding) or through mapping item pairs to category specific compatibility space (pair encoding). Bi-LSTM (coarse) performing slightly better than FHN (coarse) and CSN (coarse) probably proves that \textit{order encoding} is a better way to encode category information than the \textit{pair encoding}, because it learns a `one-many' compatibility instead of `one-one' compatibility. 
FHN (coarse) performs worse than CSN (coarse) mainly because FHN uses binary codes instead of continuous codes, which limit the encoding of information into the feature.
CSN (fine) and FHN (fine) performs better than Bi-LSTM (coarse) because they encode the fine category information into the model, which provides more detailed and enriched information.  The reason that MCAN performs better than Bi-LSTM, FHN, and CSN is because it learns the tuple features instead of only item features. 


Tuple triplet learning is helpful to better tighten the tuple compatibility space. Thus, it introduces some improvements over variants without tuple triplet learning.

One interesting observation is that MAC has more improvements over other models on IQON dataset than on Polyvore dataset. For example, MAC + Triplet (fine) achieves $6.0\%$, $9.0\%$, and $12.0\%$ improvements over CSN (fine) on IQON dataset for easy, semi-hard and hard tasks respectively. While on Polyvore dataset, the improvements drop to $0\%$, $5.0\%$, and $9.0\%$. This can be attributed to the different characteristics of the datasets. In IQON, we make sure that an outfit does not contain items of the same category, which is better fit for the practical application scenarios.

\begin{table*}[!t]
\caption{Comparison of different methods on Polyvore Dataset on the FITB and compatibility prediction tasks, with tasks created using easy, semi-hard and hard negative sampling methods. Results are provided for both ordered/unordered cases on the left and right of `/'. }
\small
\centering
\begin{tabular}{cccccccl}
\toprule
\multirow{2}{*}{Method} & \multicolumn{3}{c}{FITB accuracy (\%)} & \multicolumn{3}{c}{Compat. AUC} \\\cline{2-7}
& easy & semi-hard & hard & easy & semi-hard & hard \\
\midrule
Siamese Net \cite{VeitKBMBB15} (none) & 52.0/52.0 &  47.6/47.6 & 44.0/44.0 & 0.85/0.85 & 0.80/0.80 & 0.72/0.72 \\
Bi-LSTM \cite{han2017learning} (coarse) & 68.6/61.9 &  66.2/60.1 & 62.7/57.3 & 0.90/0.83 & 0.84/0.78 & 0.77/0.71 \\
FHN~\cite{Lu_2019_CVPR} (coarse) & 65.2/65.2 & 63.5/63.5 & 59.4/59.4 & 0.87/0.87 & 0.82/0.82 & 0.74/0.74 \\
FHN~\cite{Lu_2019_CVPR} (fine) & 83.3/83.3 & 78.5/78.5 & 75.4/75.4 & 0.92/0.92 & 0.86/0.86 & 0.80/0.80 \\
CSN \cite{VasilevaECCV18FasionCompatibility} (coarse) & 67.3/67.3 & 65.9/65.9 & 61.2/61.2 & 0.89/0.89 & 0.83/0.83 & 0.77/0.77 \\
CSN \cite{VasilevaECCV18FasionCompatibility} (fine) & 86.2/86.2 & 80.3/80.3 & 77.2/77.2 & 0.98/0.98 & 0.90/0.90 & 0.82/0.82 \\
\hline
MAN (coarse) & 81.8/79.4 & 79.2/76.0 & 77.3/74.2 & 0.90/0.85 & 0.88/0.83 & 0.81/0.77 \\
MAN (fine) & 86.1/83.3 & 83.3/80.7 & 80.6/77.9 & 0.94/0.90 & 0.91/0.88 & 0.85/0.81 \\
\hline
MCAN (coarse) & 83.6/83.6 & 81.0/81.0 & 79.1/79.0 & 0.93/0.93 & 0.90/0.89 & 0.85/0.84 \\
MCAN (fine) & 87.4/87.4 & 85.8/85.6 & 83.0/82.7 & 0.96/0.95 & 0.94/0.93 & 0.88/0.86 \\
MCAN + Triplet (coarse) & 85.4/85.3 & 82.3/82.2 & 80.2/80.0 & 0.95/0.94 &  0.92/0.91 & 0.87/0.85 \\
MCAN + Triplet (fine) & \textbf{89.8}/\textbf{89.8} & \textbf{86.5}/\textbf{86.4} & \textbf{84.3}/\textbf{84.1} & \textbf{0.98}/\textbf{0.98} & \textbf{0.95}/\textbf{0.94} & \textbf{0.91}/\textbf{0.88} \\
\bottomrule
\end{tabular}
\label{tab:polyvore}
\end{table*}

\subsection{Fill-In-The-Blank Outfit Recommendation}
In the FITB task,
one has to choose the most compatible item for an outfit from multiple choices. One application example is when a customer wants to choose a pair of boots to match his jacket and pants. We use all outfits in the Polyvore and IQON test datasets to create the FITB tasks. For each outfit, we randomly select one item as the groundtruth and replace it with a blank to make a question. Then we select 3 other negative items along with this groundtruth to obtain an answer set with 4 choices.  We assume that items randomly selected from other outfits make less compatible outfits than the groundtruth. We use FITB accuracy as the evaluation metric. A model is more efficient if it can answer more questions correctly. To answer the question, we fill each answer in the blank and select the one based on the following objective:
\begin{equation}
    \hat{\mathbf{x}}_a = \argmax_{\mathbf{x}_a \in \mathbf{A}} P\big(\mathbf{x}_{a}\vert(\mathbf{x}_1, \mathbf{c}_1), \ldots, (\mathbf{x}_{i}, \mathbf{c}_{i}),\mathbf{c}_{i+1}\big)
\label{eq:prob2}
\end{equation}
where $\mathbf{A}$ is the answer set. The candidate having the highest probability is selected as the final answer.

The left parts of Table \ref{tab:iqon} and \ref{tab:polyvore} show the results of our method compared with other baselines. 
Similar to  the results from the outfit compatibility studies, MCAN performs the best. With the consideration of both coarse and fine categories, MCAN is more flexible and can better adapt to the customer needs. 
For example, suppose a customer wants a pair of boots and a school bag to match his sweater and jeans, MCAN can guarantee the next recommended items to be sweater and jeans, while Bi-LSTM cannot because it is only trained on coarse categories. 
Similarly, items with fine categories can be recommended using CSN (fine) or FHN (fine) but not CSN (corase) or FHN (coarse), which means CSN and FHN have to be re-trained to meet different needs. 
MCAN can work in both cases.

\subsection{Evaluation on Random Ordering}
\label{ro}
Sequential models like BiLSTM \cite{han2017learning,Takuma} require that the items in an outfit be organized in a fixed order. In Table~\ref{tab:iqon} and ~\ref{tab:polyvore}, after shuffling the outfit items in the test test of IQON and Polyvore datasets, the compatibility AUC drops for all tasks. 
Similar performance impacts are observed for MAN (coarse) and MAN (fine), where no CPLs are used. Thus, sequential models have better performance if the test data keeps the same order as the training data. However, if order varies, the performance degrades. We solve this problem by adding a category prediction layer (CPL), which helps to align the item sequence with the category sequence. As category information is easily obtained in real world applications, aligning item sequence with category sequence greatly improves the performance of recommendation systems and makes MCAN `order-free'.

\subsection{Controllable Diverse Recommendation}
\label{sec:cdr}
\begin{figure}
\centering
\includegraphics[width=0.45\textwidth]{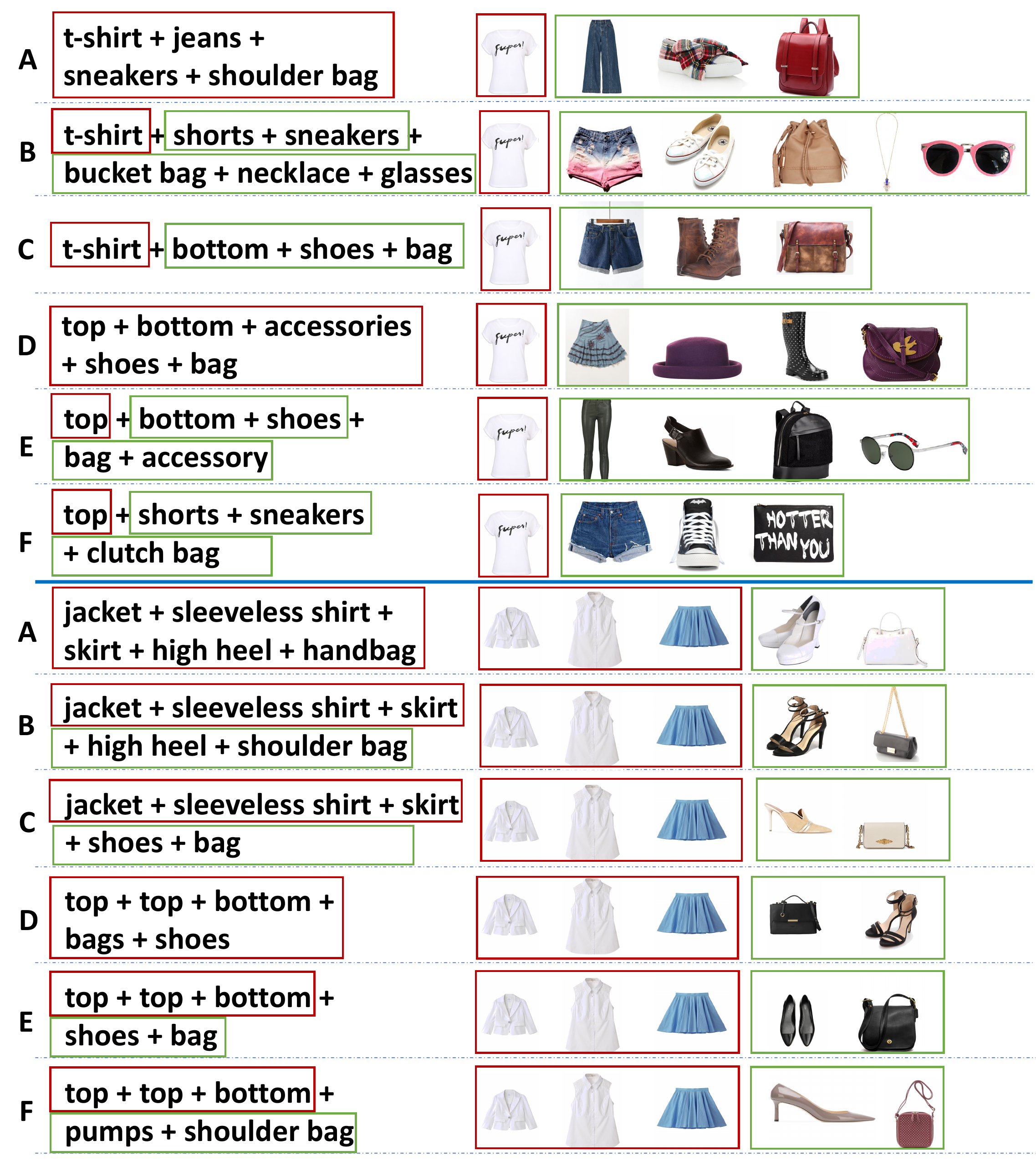}
\caption{\textbf{Controllable and diverse recommendation given query tuples.} Each row contains a recommended outfit where query tuples (partial outfit) are indicated by red boxes and predicted items and categories are bounded by green boxes. The outfit above the blue line is generated from Polyvore dataset, while the one below the blue line is generated from IQON dataset.}
\label{fig:generated}
\vspace{-0.2in}
\end{figure}

MCAN can be applied to complete six different recommendation tasks, shown as $A$, $B$, $C$, $D$, $E$, and $F$ in Figure~\ref{fig:generated}.
In the case $A$, MCAN can recommend items to fit the fine categories provided by customers, which are listed in a red box. This demonstrates that MCAN is able to generate controllable recommendations by feeding in the category requirements. Case $B$ signifies that MCAN can decode fine tuple sequence to complete the outfit, with only partial category sequence provided. The items or categories provided in the green box are predicted by MCAN, while those in the red box are provided by customers. In case $C$, tuples with fine categories are provided as the input. MCAN recommends a sequence of tuples with coarse categories, making a `mixed' recommendation.
$D$, $E$ and $F$ are similar to $A$, $B$ and $C$, but with coarse categories. MCAN can generate more controllable and diverse recommendations than BiLSTM or CSN, which recommends by beam-searching or replacing items in existing outfits with similar ones.

\subsection{Replacing Target Item with Alternatives}
\begin{figure}[!htb]
\centering
\includegraphics[width=0.3\textwidth]{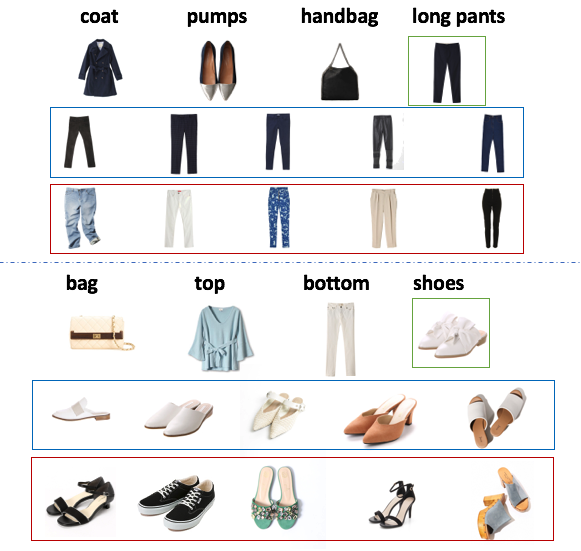}
\caption{Two examples for replacing the target item with alternatives. Top rows represent valid outfits in the dataset. In each example, the one that is highlighted in green is a randomly selected item to be replaced.
The other items on the first row remain unchanged. Second rows (highlighted in blue) display top $5$ alternatives recommended by MCAN. Bottom rows (highlighted in red) show $5$ random selection of alternatives of the same category.}
\label{fig:alternative}
\vspace{-0.2in}
\end{figure}

Given a set of fashion items, a recommendation system has to suggest several highly compatible items (the alternatives to the target one) to the customer to choose from.
To demonstrate this, we randomly select an item from the valid outfit curated by fashion designers in the dataset.
We then generate top-5 compatible items using our MCAN model.
We also randomly select 5 alternatives of the same category as the selected one.
We show the comparison results in Fig.~\ref{fig:alternative}.
We demonstrate that MCAN can recommend items with the similar styles and colors to the target one.
This demonstrates that MCAN can use both aesthetics and categories to provide a better recommendation.

\section{Conclusion}
\label{conclusion}
In this paper, we propose a Mixed Category Attention Net (MCAN), a novel and effective model for learning the tuple compatibility of outfits and conducting controllable and diverse outfit recommendation. A customer can flexibly provide either coarse or fine type of items in an outfit, based on which MCAN will determine the compatibility of the outfit or recommend remaining items to enrich the outfit. To achieve the goal, MACN incorporates both coarse and fine categories into the compatibility learning and propose tuple triplet learning to tighten the compatibility space. We also propose semi-hard and hard negative sampling methods, which not only helps improve the tuple triplet learning but also allows for creating tasks of different difficulties to test the performance of different models.
MCAN achieves the state-of-the-art performances on compatibility prediction and FITB tasks regardless of the order of items in an outfit. 

\section{Acknowledgements}
This work is supported in part by the National Science Foundation under Grants NSF ECCS 2030063 and NSF CCF 2007313.

\bibliographystyle{ACM-Reference-Format}
\bibliography{acmart}

\end{document}